%% file: main.tex
\documentclass{article} 
\usepackage{analemma,times}

\input{math_commands.tex}

\usepackage{hyperref}
\usepackage{url}
\usepackage{graphicx}
\usepackage{subfigure}
\usepackage{booktabs}
\usepackage{adjustbox}
\usepackage{amsmath}
\usepackage{amssymb}
\usepackage{mathtools}
\usepackage{amsthm}

\usepackage{multirow}
\usepackage{color}
\usepackage{colortbl}
\usepackage[capitalize,noabbrev]{cleveref}
\usepackage{xspace}

\theoremstyle{plain}

\theoremstyle{definition}

\theoremstyle{remark}

\graphicspath{{figures/}} 

\title{FARS: A Fully Automated Research System Deployed at Scale}

\newcommand{\farsreviewers}{%
Bobo Li\textsuperscript{1}, Changze Lv\textsuperscript{2}, Cheng Xu\textsuperscript{3}, Chengsong Huang\textsuperscript{4}, Chunyang Li\textsuperscript{5}, Dizhan Xue\textsuperscript{6},
Hao Bai\textsuperscript{7}, Haodong Duan\textsuperscript{8}, Hengquan Guo\textsuperscript{9}, Hongyang He\textsuperscript{10}, Hongyi Chen\textsuperscript{11}, Hui Shen\textsuperscript{12},
Jiahao Yuan\textsuperscript{13}, Jiankai Sun\textsuperscript{14}, Jikang Cheng\textsuperscript{15}, Jinfeng Xu\textsuperscript{16}, Jingqi Tong\textsuperscript{2,17}, Jingye Chen\textsuperscript{5},
Jinxiu Liu\textsuperscript{18}, Jixuan Leng\textsuperscript{11}, Junchi Yu\textsuperscript{19}, Kaixun Jiang\textsuperscript{2}, Kun Xiang\textsuperscript{20}, Kunpeng Yao\textsuperscript{21},
Lang Feng\textsuperscript{22}, Liangqi Yuan\textsuperscript{23}, Longsen Gao\textsuperscript{24}, Meng Li\textsuperscript{25}, Qi Jia\textsuperscript{26}, Qiushi Sun\textsuperscript{16},
Shengyuan Ding\textsuperscript{2}, Shizhan Gong\textsuperscript{27}, Siru Zhong\textsuperscript{28}, Terry Jingchen Zhang\textsuperscript{29}, Tianle Gu\textsuperscript{30}, Tianyi Liang\textsuperscript{17,31},
Weijie Liu\textsuperscript{15}, Weikai Yang\textsuperscript{28}, Weizhi Fei\textsuperscript{30}, Xiangkun Hu\textsuperscript{32}, Xiangyang Liu\textsuperscript{32}, Xin Wang\textsuperscript{33},
Xinpeng Liu\textsuperscript{34}, Xuanwen Ding\textsuperscript{2}, Yihong Tang\textsuperscript{35,36}, Yuanli Wang\textsuperscript{37}, Yukun Jiang\textsuperscript{38}, Yuming Yang\textsuperscript{2},
Zhengbao He\textsuperscript{34}, Zhikai Chen\textsuperscript{39}, Zhikun Xu\textsuperscript{40}, Zhuang Li\textsuperscript{41}, Zihao Huang\textsuperscript{8}, Anonymous,
Anonymous%
}

\newcommand{\farsrevieweraffiliations}{%
\textsuperscript{1}National University of Singapore, \textsuperscript{2}Fudan University, \textsuperscript{3}University College Dublin,
\textsuperscript{4}Washington University in St. Louis, \textsuperscript{5}The Hong Kong University of Science and Technology, \textsuperscript{6}Institute of Automation, Chinese Academy of Sciences,
\textsuperscript{7}University of Illinois at Urbana-Champaign, \textsuperscript{8}ByteDance, \textsuperscript{9}ShanghaiTech University,
\textsuperscript{10}University of Warwick, \textsuperscript{11}Carnegie Mellon University, \textsuperscript{12}University of Michigan, Ann Arbor,
\textsuperscript{13}East China Normal University, \textsuperscript{14}Stanford University, \textsuperscript{15}Tencent,
\textsuperscript{16}The University of Hong Kong, \textsuperscript{17}Shanghai Innovation Institute, \textsuperscript{18}Nex-AGI Team,
\textsuperscript{19}University of Oxford, \textsuperscript{20}Sun Yat-sen University, \textsuperscript{21}University of Leeds,
\textsuperscript{22}Nanyang Technological University, \textsuperscript{23}Purdue University, \textsuperscript{24}University of New Mexico,
\textsuperscript{25}Nanjing University, \textsuperscript{26}Shanghai Artificial Intelligence Laboratory, \textsuperscript{27}The Chinese University of Hong Kong,
\textsuperscript{28}The Hong Kong University of Science and Technology (Guangzhou), \textsuperscript{29}Vector Institute, \textsuperscript{30}Tsinghua University,
\textsuperscript{31}OpenMOSS, \textsuperscript{32}Analemma, \textsuperscript{33}The Ohio State University,
\textsuperscript{34}Shanghai Jiao Tong University, \textsuperscript{35}McGill University, \textsuperscript{36}ServiceNow AI Research,
\textsuperscript{37}Boston University, \textsuperscript{38}CISPA - Helmholtz-Zentrum f\"{u}r Informationssicherheit gGmbH, \textsuperscript{39}Michigan State University,
\textsuperscript{40}Arizona State University, \textsuperscript{41}RMIT University%
}

\author{\normalfont
\begin{minipage}{0.95\textwidth}
\raggedright
\textbf{Authors}\\[1.0ex]
Qiong Tang\textsuperscript{\textdagger}, Tianxiang Sun\textsuperscript{\textdagger}, Xiangkun Hu\textsuperscript{\textdagger}, Xiangyang Liu\textsuperscript{\textdagger}, Yiran Chen\textsuperscript{\textdagger}, Yunfan Shao\textsuperscript{\textdagger}\\
Analemma\\
\texttt{\{qtang,txsun,xkhu,xyliu,yrchen,yfshao\}@analemma.ai}\\[2.8ex]
\textbf{Reviewers}\textsuperscript{\(\ddagger\)}\\[1.0ex]
\farsreviewers\\[0.8ex]
{\footnotesize \farsrevieweraffiliations}
\end{minipage}
}

\analemmafinalcopy 
\begin{document}

\maketitle
\begingroup
\renewcommand{\thefootnote}{\fnsymbol{footnote}}
\footnotetext[2]{Equal contribution; human authors listed in alphabetical order by first name.}
\footnotetext[3]{Named reviewers are listed alphabetically.}
\endgroup

\newpage
\begin{abstract}
\input{sections/abstract}
\end{abstract}

\input{sections/introduction}

\input{sections/related_work}

\input{sections/method}

\input{sections/deployment}

\input{sections/experiments}

\input{sections/comparison}

\input{sections/conclusion}

\bibliography{analemma}
\bibliographystyle{analemma}

\appendix
\input{sections/appendix}

\end{document}

%% file: math_commands.tex
\usepackage{amsmath,amsfonts,bm}

\def\eqref#1{equation~\ref{#1}}

\def\1{\bm{1}}

\DeclareMathAlphabet{\mathsfit}{\encodingdefault}{\sfdefault}{m}{sl}
\SetMathAlphabet{\mathsfit}{bold}{\encodingdefault}{\sfdefault}{bx}{n}



%% file: sections/abstract.tex
Recent automated research systems show that language-model agents can generate hypotheses, run experiments, and write complete manuscripts, but most evidence still comes from selected examples, human-framed topics, or a few pre-defined research tasks. We present FARS (Fully Automated Research System), a fully automated AI-for-AI research system designed to operate across research topics at scale. FARS autonomously generates and advances projects through ideation, planning, experimentation, and writing, using stage-specific agents coordinated through a shared workspace that records proposals, code, logs, results, and manuscripts. In its first public deployment, FARS produced 166 complete research papers spanning 67 fine-grained AI/ML topics while preserving intermediate artifacts as an auditable corpus rather than a curated set of successes. We evaluate this corpus with 282 structured reviews from volunteer reviewers covering 140 papers, including overall ratings, sub-scores, integrity checks, and LLM-use disclosure. The reviews indicate that FARS can produce review-worthy and occasionally strong AI/ML research artifacts in a large-scale public deployment, while also exposing recurring failure modes in narrow experimental scope, methodological limitations, and integrity issues.

%% file: sections/introduction.tex
\section{Introduction}
\label{sec:intro}

Recent progress in autonomous research systems suggests that language-model agents can now perform substantial portions of the scientific workflow, including literature review, hypothesis generation, experiment execution, reviewing, and paper writing~\citep{asai2024openscholarsynthesizingscientificliterature,baek2025researchagentiterativeresearchidea,schmidgall2025agentlaboratoryusingllm,li2025mlrcopilot}. Systems such as The AI Scientist~\citep{lu2024aiscientistfullyautomated}, CycleResearcher~\citep{weng2025cycleresearcher}, Zochi~\citep{zochi2025}, AI Scientist v2~\citep{yamada2025aiscientistv2workshoplevelautomated}, AIGS~\citep{liu2024aigsgeneratingscienceaipowered}, and AI-Researcher~\citep{tang2025airesearcher} demonstrate increasingly complete research pipelines, while DeepScientist~\citep{weng2025deepscientist} shows that autonomous search can make measurable progress on human-defined frontier tasks. However, the existing evidence base remains limited in scope, relying largely on selected demonstrations, human-framed topics, user-provided references, narrowly defined state-of-the-art improvement targets and benchmark tasks. These settings provide useful signals about what autonomous research agents can do, but they do not yet show how such systems behave when they run continuously, select and execute work at scale, and produce large volumes of outputs. This leaves open a central systems question: what happens when a fully automated AI research system is deployed continuously at scale, and how should its outputs be evaluated?

We introduce FARS, an end-to-end AI research system that operates at scale. It is designed to autonomously perform the complete research workflow without human intervention during execution. The guiding principle of FARS is to efficiently and reliably expand the frontier of knowledge. Under this view, the unit of automated research is a focused contribution: a clearly articulated hypothesis paired with an empirical or theoretical validation attempt. Narrowly scoped findings and negative results remain valuable as long as they are observable, well motivated, and verifiable. In its current instantiation, FARS is applied to AI research, positioning it within an emerging AI-for-AI research paradigm. AI research provides a practical deployment domain for autonomous research systems: many hypotheses can be operationalized as executable code and evaluated against shared benchmarks, while advances in AI have high intrinsic value and can also accelerate research in other fields.

FARS organizes research into four sequential stages, namely Ideation, Planning, Experiment, and Writing, coordinated through a shared workspace that serves both as persistent project memory and as an auditable artifact store. In its first continuous public deployment, FARS produced 166 complete papers spanning 67 fine-grained AI/ML topics. This scale contrasts with prior end-to-end systems, whose evaluations typically rely on selected papers, workshop submissions, human-scoped benchmark tasks, or a small number of pre-defined frontier tasks. Our deployment of FARS was designed as a live, large-scale experiment: FARS outputs vary across topics, methods, and quality, making evaluation through a handful of selected examples necessarily limited. We therefore retained successes, weak outputs, negative results, execution artifacts, and failure modes for inspection rather than presenting a curated sample, while external dissemination required human review and explicit labeling.

We evaluate the papers produced during this public deployment based on 282 structured reviews from carefully recruited volunteer reviewers covering 140 generated papers.\footnote{The review cycle ran from March 21 to April 12, 2026; 140 of the 166 generated papers received at least one completed review, while the remaining 26 were not reviewed and are excluded from the statistics.} The reviews include overall ratings, scores for soundness, presentation, and contribution, reviewer confidence, integrity checks, and disclosure of LLM-assisted review. The results reveal substantial quality variation: FARS can produce review-worthy and sometimes strong research artifacts at a public-deployment scale, while recurring weaknesses include narrow experimental scope, methodological limitations, and integrity issues. These findings suggest that scalable automated research is already feasible, while the reliability of its scientific contribution is still bounded by the substantive value of the contribution, the sufficiency of experimentation, and the faithfulness of writing. Our contributions are as follows:

\begin{itemize}
    \item A fully automated research system, FARS, that turns research directions into complete papers through dedicated Ideation, Planning, Experiment, and Writing agents coordinated by a shared auditable workspace.
    \item A public deployment of FARS that produced 166 complete papers spanning 67 fine-grained AI/ML topics while preserving an auditable corpus of intermediate artifacts, including hypotheses, plans, code, logs, results, and manuscripts.
    \item A large-scale human review of deployment outputs, measuring paper quality, recurring failure modes, and the gap between scalable generation and reliable scientific contribution.
\end{itemize}

%% file: sections/related_work.tex
\section{Related Work}

\paragraph{Autonomous Research Systems.}
Autonomous research systems have progressed from assisting isolated research tasks toward automating larger portions of the scientific workflow. The AI Scientist~\citep{lu2024aiscientistfullyautomated} introduced an end-to-end pipeline for idea generation, code generation, experimentation, visualization, paper writing, and automated review, while AI Scientist v2~\citep{yamada2025aiscientistv2workshoplevelautomated} reduced template dependence and demonstrated workshop-level AI-generated manuscripts. Subsequent systems extend this direction through benchmark-based autonomous scientific innovation~\citep{tang2025airesearcher}, goal-oriented frontier improvement~\citep{weng2025deepscientist}, evidence-grounded claim verification~\citep{meng2026scientistonechainofevidence}, automated research-review cycles~\citep{weng2025cycleresearcher}, falsification-driven discovery~\citep{liu2024aigsgeneratingscienceaipowered}, and human-verified high-autonomy scientific discovery~\citep{zochi2025}. Collectively, these works show that agents can automate substantial parts of the research lifecycle, but they are typically evaluated within human-scoped domains, tasks, templates, benchmark corpora, starting methods, or submission and verification procedures. FARS differs by treating automated AI research as a deployed system for fully autonomous research project progression: from idea generation through planning, experimentation, and paper writing, each project proceeds without human intervention during execution.

\paragraph{Large-Scale Automated Research.}
Scale in automated research can mean large search spaces, many experiments, or many generated artifacts. DeepScientist~\citep{weng2025deepscientist} reports thousands of generated ideas and over a thousand experimental validations, but concentrates this search on three frontier tasks initialized from human SOTA methods. AI-Researcher~\citep{tang2025airesearcher} evaluates across 22 curated benchmark papers and open-ended variants, but the benchmark is constructed around selected human-authored papers, references, and datasets. AutoSOTA~\citep{li2026autosota} maps recent top-tier AI papers to executable repositories and optimizes them into improved models, while CodeScientist~\citep{jansen2025codescientist} runs hundreds of code-based experiments, returning 19 candidate discoveries of which 6 pass external and internal validation. Robin~\citep{ghareeb2026robin} demonstrates an iterative lab-in-the-loop discovery workflow in experimental biology. These systems demonstrate scale in search, optimization, and empirical evaluation within structured settings. FARS instead examines scale through a public deployment of the idea-to-paper pipeline, combining complete paper generation with preserved intermediate artifacts.

\paragraph{Evaluation of AI-Generated Research.}
Evaluation remains difficult because research quality depends on soundness, contribution, evidence, and integrity, not only on fluent presentation. LLM-as-a-Judge methods~\citep{zheng2023judgingllmasajudgemtbenchchatbot,gu2025surveyllmasajudge} scale cheaply, but they may miss field-specific weaknesses; related work studies overlap between LLM feedback and human review~\citep{liang2023largelanguagemodelsprovide} and develops standardized automated review protocols~\citep{yu2024automatedpeerreviewingpaper}. Idea quality is also insufficient on its own: LLM-generated ideas can be rated as novel while remaining less feasible~\citep{si2024llmsgeneratenovelresearch}, and this ideation--execution gap widens when ideas are implemented~\citep{si2025ideationexecutiongap}. Benchmarks such as MLAgentBench~\citep{huang2024mlagentbenchevaluatinglanguageagents}, EXP-Bench~\citep{kon2025expbenchaiconductai}, PaperBench~\citep{starace2025paperbench}, and MLR-Bench~\citep{chen2025mlrbenchevaluatingaiagents} evaluate experimentation, replication, and open-ended ML research, finding that agents still struggle with execution and validation. Recent assurance-oriented systems further emphasize claim auditing, verifiable reporting, and integrity safeguards~\citep{yang2026arisautonomousresearchadversarial,liu2026autoresearchclawselfreinforcingautonomousresearch,gupta2025plagiarism,zhu2025safescientist,resnik2026autonomous}. FARS instead evaluates complete papers produced by a deployed autonomous research system using structured human reviews, sub-scores, integrity checks, LLM-use disclosure, and preserved artifacts for inspection.

%% file: sections/method.tex
\section{FARS}
\label{sec:method}

\begin{figure}[t]
\centering
\includegraphics[width=\textwidth]{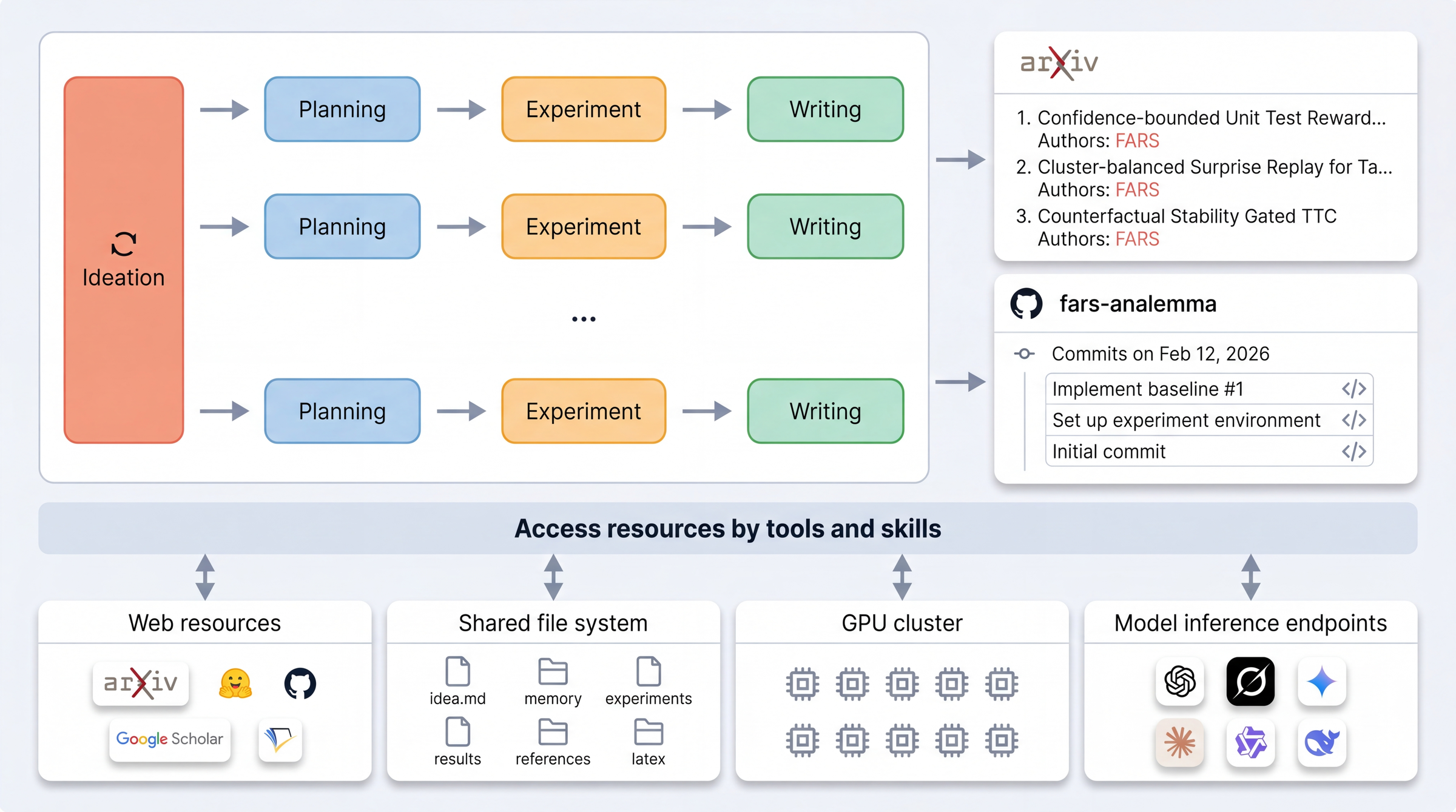}
\caption{Overview of the FARS architecture. The Ideation stage generates multiple research proposals through iterative refinement, and accepted proposals proceed through Planning, Experiment, and Writing stages. Multiple projects can run independently while sharing a common pipeline structure. Completed manuscripts are paired with code repositories and supporting artifacts. All stages coordinate through a shared workspace, which serves as persistent project memory and an artifact store for downstream inspection; agents also access supplementary tools and infrastructure, including network resources, curated skills, a GPU cluster with 160 NVIDIA GPUs, and model inference endpoints.}
\label{fig:framework}
\end{figure}

FARS is a multi-agent research system that turns open-ended research directions into complete hypothesis-and-validation papers. It consists of four specialized stages, namely Ideation, Planning, Experiment, and Writing that sequentially transform candidate research hypotheses into executable plans, experimental evidence, and manuscripts. All stages coordinate through a shared workspace, which serves as both project memory and an artifact store for downstream inspection. Figure~\ref{fig:framework} provides an overview of the pipeline.

\subsection{Design Principles}

FARS follows three design principles. First, the unit of production is a focused research contribution rather than a maximally polished conference paper. A project should state a hypothesis and make an empirical or theoretical validation attempt. FARS does not treat a negative result as a failed paper by default: when an experiment clearly tests the hypothesis and the limitations are explicit, the trajectory can become a negative-result manuscript rather than being suppressed. Second, the system is artifact-centered: agents exchange stage outputs through a shared workspace rather than hidden state alone, leaving both successful and failed trajectories inspectable. These artifacts provide the interface between stages and the evidence base for downstream inspection. Boundary checks validate stage outputs and trigger retry or feedback when supported. Third, FARS separates ideation, planning, experimentation, and writing into stages with clear inputs, outputs, and checks, allowing many projects to run independently while preserving a common structure for downstream review.

\subsection{Ideation Stage}



Ideation is the first stage of FARS, responsible for turning open-ended research directions into concrete, executable proposals. In the live deployment, it operates without any interactive user input: rather than responding to individual requests, the agent is seeded with nine suggested research topics and is encouraged to explore freely beyond them. From this starting point, it autonomously performs literature review, direction exploration, proposal drafting, and quality assessment. Each proposal that clears automated review is organized around a single, well-scoped hypothesis, covering motivation, approach, related work, experimental design and success criteria. The generated research proposal serves as the input to the Planning stage and is also carried over to other downstream stages of Experiment and Writing.

Internally, Ideation is a collaboration among four specialized agents. A Lead Agent that orchestrates the process—searching open-access papers and public code repositories, curating a three-tier knowledge base (raw text, summaries, and a synthesized domain survey), identifying research gaps, and authoring the proposal; a Summarization Agent that distills the methods, results, and limitations of retrieved papers; a Peer Discussion Agent that adversarially pre-reviews candidate directions to prune weak ideas early; and an Evaluation Agent that conducts the  accept/revise/reject judgment and provides revision feedback to the Lead Agent. Together, they drive a six-stage pipeline in an agentic manner: literature review, generation of candidate directions, peer discussion and filtering, proposal drafting, an iterative review-and-revision loop, and finalization.

Since every accepted proposal is forwarded downstream to produce a paper, the live system runs end-to-end: a proposal is admitted only once it clears an automated quality gate, with a circuit breaker that selects the best available draft after repeated failures to prevent unbounded iteration. Candidate directions are further constrained to remain within the AI/ML domain and to yield fully automated experiments that require no human annotation and stay within a fixed compute budget, with detailed planning and implementation deferred to downstream modules.

\subsection{Planning Stage}



The Planning stage is tightly coupled with the Experiment stage, and its design is driven by the demands of long-horizon research execution. Unlike the software-engineering tasks at which coding agents typically excel, a complete research study may take hours or even days to run, yet FARS is intended to carry it out without human intervention. On such long-horizon tasks, current models exhibit two failure modes that are fatal for research: they tend to attempt the entire task in a single pass, giving insufficient thought to each sub-task, and they often declare a task complete after only a shallow attempt~\citep{anthropic2025effectiveharnesses}. Research projects, however, demand fine-grained control over every detail and a complete, reliable record of experimental procedures and results. To mitigate these failure modes, the Planning stage does not hand the Experiment stage a free-form goal; instead, it converts the accepted proposal into a machine-readable experiment contract that decomposes the study into discrete, ordered units of work.

Concretely, the planning agent first reads the upstream proposal to understand its hypothesis, candidate method design, baselines to compare against, and target benchmarks, and then surveys the experimental setups of related work as a reference for what a rigorous study in this area should contain. Guided by this context, it constructs a structured experiment plan using a two-level hierarchical structure: each entry is an experiment item assigned to one of five task categories, and every item is further decomposed into ordered steps with concrete, executable instructions. The Experiment stage then consumes this structure and completes the steps one item at a time, so that each sub-task receives dedicated attention rather than being rushed in a single sweep.

The five task categories are environment configuration, baseline experiments, main experiments, effectiveness evaluation, and analysis experiments. Together, they encode the intended evidential structure of the study: the environment is prepared first, baselines establish comparison points, main experiments test the proposed method, effectiveness evaluation determines whether the evidence supports continuing, and analysis experiments deepen the interpretation only after the main claim has passed. Automated validation checks the JSON format, required fields, category names, and ordering constraints. Failed validation produces targeted feedback for regeneration, so the plan handed to the Experiment stage is both executable and faithful to the proposal's experimental logic.

\subsection{Experiment Stage}

The Experiment stage validates whether the hypothesis in the proposal can be realized. It must faithfully execute every task in the validated plan and organize the results so that the Writing stage can reference them programmatically. The central design constraint is that, for autonomous research, execution success alone is not sufficient: the system must preserve enough context for the external observer to determine what was run, what was measured, and whether the results support the claim. This auditability requirement shapes every mechanism in the stage.

\paragraph{The Experiment Loop}
An autonomous agent given a multi-task experiment plan may deviate in subtle ways, such as adding unrequested experiments, modifying success criteria, or declaring success on incomplete results. Because the agent controls both execution and self-reporting, such deviations are difficult to detect after the fact. The stage therefore separates orchestration from execution, in which the harness decomposes the validated plan into sequential experiment items and executes them one at a time. Between items, the harness maintains a progress record and accumulates result summaries so that each subsequent agent understands what has been accomplished and can build on prior findings. An effectiveness evaluation gates the transition from main experiments to analysis, ensuring that compute is not spent investigating unsupported hypotheses while preserving negative outcomes as first-class results~\citep{chen2025mlrbenchevaluatingaiagents}. Within each item, execution follows a structured loop. The experiment agent first produces an execution plan by examining the current environment, searching for available models and datasets and activating relevant skills. An independent review agent evaluates the fidelity of the execution plan to the proposal and iterates until approved. The experiment agent then carries out the approved work with full tool permissions. Upon completion, dual verification combines the agent's semantic self-assessment with deterministic harness checks on output completeness, and results are written to a pre-defined location as experimental evidence which can be used in the paper writing.

\paragraph{Layered Environments and Curated Skills}
ML software stacks are fragile due to tightly coupled CUDA builds, compiled extensions, and interdependent library versions, and an agent that must assemble them from scratch wastes most of its effort on setup failures rather than experimentation. FARS addresses this with a layered environment: a pre-configured base provides heavy dependencies (PyTorch, vLLM, flash-attention, Transformers, etc.) that the agent cannot modify, while the agent works inside an isolated environment where it can freely install additional packages. Pre-trained model weights and common datasets are cached on shared storage, eliminating repeated downloads. A related challenge is that language models possess broad but shallow knowledge of ML tooling and conventions: they know the concepts behind distributed training or evaluation frameworks but frequently use deprecated APIs or miss known pitfalls. To bridge this gap, FARS maintains a curated skill library covering 19 categories of ML research practice, organized as structured reference files with verified code examples and troubleshooting guides. Rather than loading all skills into context at once, the agent is given only a category-level index and retrieves individual skills on demand, first scanning the index, then reading a brief description to judge relevance, and loading full content only when needed.

\paragraph{Managed Compute and Unified Model Access}
Experiments require GPU-accelerated training and access to diverse model APIs for tasks such as LLM-based evaluation, synthetic data generation, and embedding computation. Exposing raw infrastructure interfaces to the agent risks uncontrolled resource allocation, orphaned processes, and operations untraceable to the plan. FARS instead provides high-abstraction tool interfaces through which the agent declares its intent (model, GPU count, training configuration) while the platform handles resource provisioning and lifecycle management. Each task is tagged with the corresponding plan item for attribution, and ownership isolation ensures agents can only access their own resources. All compute is automatically reclaimed upon completion or interruption. For model API access, a unified service proxy presents multiple open-source and proprietary providers behind a single compatible endpoint, supporting chat completion, embedding, reranking, and image generation with centralized authentication and usage tracking, so that the agent focuses on experimental logic rather than provider-specific integration.

\paragraph{Checkpoint-Based Fault Tolerance}
A full experiment may span hours or days, so the stage is designed for robustness under long, unattended operation. Each task's outputs are committed to version control immediately upon completion, so that finished results survive even if a later task or the run itself fails. Checkpoint-based recovery resumes from the last unfinished item, preserving the agent's full context for continuity. Automated scanning runs before each commit to prevent credential leakage, and a termination handler stops all associated compute tasks to prevent orphaned processes.

\subsection{Writing Stage}

The Writing stage transforms the complete upstream project trajectory into an academic paper. FARS separates evidence analysis from manuscript drafting: the \textit{Analysis Agent} determines what can be claimed from the proposal, plan, experiments, and related work, while the \textit{Writing Agent} turns this evidence plan into a polished manuscript. This decomposition is intended to prevent claims from being introduced opportunistically during prose generation and to ensure that writing remains grounded in the artifacts produced by earlier stages.

Before drafting begins, the upstream materials are audited and organized into a paper blueprint. The blueprint is a structured intermediate representation that links each central claim to its supporting experiments, source artifacts, candidate figures and tables, and relevant literature. It also defines the section-level writing plan used by the downstream drafting process. In addition to organizing evidence, the Analysis Agent prepares visual materials: analytical plots are generated by executing plotting scripts over saved experimental outputs, while methodological diagrams are synthesized from detailed figure descriptions and refined with vision-language model feedback. In our implementation, the diagram synthesis step is instantiated with NanoBanana-Pro~\citep{google2025nanobananapro}.

Guided by the blueprint, each section is decomposed into concrete writing tasks that inherit the blueprint's evidence links, keeping drafting tied to the planned claims and supporting artifacts. The agent drafts the LaTeX manuscript section by section and then refines it through automatic compilation feedback, citation and reference checks, figure and table checks, numerical verification against source artifacts, and paper-level review for unsupported or overstated claims. By placing evidence organization before prose generation and tracing numerical claims, tables, and plots back to source artifacts during refinement, the Writing stage reduces hallucinated results and unsupported interpretations. The final paper is therefore maintained together with the proposal, plan, code, logs, results, and supporting evidence needed to inspect how its claims were produced.

%% file: sections/deployment.tex
\section{Public Deployment}
\label{sec:deployment}

The first public instantiation of FARS produced 166 complete research papers. The live deployment ran for 417 hours, consumed 21.6 billion model tokens, and reported a total cost of approximately \(\$186{,}000\) (including GPU cluster usage fees and token consumption costs). Amortized over the completed corpus, this corresponds to 2.51 deployment hours, approximately 130 million model tokens, and about \(\$1{,}120\) per paper. The generated papers were made available through a public deployment page.\footnote{\url{https://analemma.ai/fars/}} The purpose of the deployment was not to present a curated set of selected successes, but to expose the output distribution of a continuously running AI-for-AI research system, including weak projects, negative results, and failure modes.

FARS was seeded with nine initial AI research directions, summarized in \Cref{tab:seed_topics}. These directions were intended to provide starting points rather than a closed taxonomy: the Ideation agent was allowed to explore beyond them during deployment.

\begin{table}[t]
\centering
\scriptsize
\caption{Initial research directions provided to FARS at the start of the public deployment.}
\label{tab:seed_topics}
\setlength{\tabcolsep}{3.5pt}
\adjustbox{max width=\textwidth}{
\begin{tabular}{p{0.29\textwidth}p{0.66\textwidth}}
\toprule
Seed topic & Description \\
\midrule
Reinforcement Learning from Verifiable Rewards & Training LLMs using reward signals from programmatically verifiable outcomes rather than learned reward models or human preferences. \\
Post-Training of Small Language Models & Post-pretraining techniques that transform base SLMs into useful, safe, and specialized systems. \\
Automated Evaluation of Frontier LLMs & Scalable evaluation methods for increasingly capable LLMs, where human evaluation becomes a development bottleneck. \\
Model Architecture Innovation Beyond Transformers & Alternatives to standard Transformers that address efficiency and scaling bottlenecks such as quadratic attention. \\
Continual Learning & Methods that allow models to learn from continuous data streams while retaining previously acquired knowledge. \\
Diffusion Language Models & Applying diffusion-model principles to discrete text generation and language modeling. \\
Memory in AI Agents & Mechanisms that let AI agents store, organize, retrieve, and use information across interactions and over time. \\
Test-Time Scaling & Techniques that improve LLM performance by allocating additional computation during inference. \\
World Models & Learned internal representations that allow agents to simulate environment dynamics, plan, and predict future states. \\
\bottomrule
\end{tabular}
}
\end{table}

The resulting 166 papers span 67 fine-grained topics rather than remaining within the seed set; Appendix~\ref{sec:appendix_deployment} provides the full topic distribution. Overall, 72 papers fall under the nine seed topics, while 94 papers occupy emergent topics discovered during autonomous exploration. The largest seed-topic clusters are Reinforcement Learning from Verifiable Rewards (13 papers), Automated Evaluation of Frontier LLMs (10), Memory in AI Agents (10), and Continual Learning (9). The largest emergent cluster is AI Safety and Alignment (13 papers), followed by Vision-Language Models (7) and Code Generation (6). This corpus is the basis for the human review study reported in \Cref{sec:experiments}.

The run was streamed publicly, and intermediate artifacts including hypothesis documents, plans, code, logs, draft sections, and final papers were committed to a dedicated GitLab namespace,\footnote{\url{https://gitlab.com/fars-a}} operated by the system itself. This artifact trace made the operation of the system observable beyond finished manuscripts: observers could inspect how each hypothesis-and-validation contribution was produced. The deployment therefore exposes both the final research outputs and the process evidence needed to audit them.

Submitting selected papers to archival venues such as arXiv was handled separately from the public deployment. While the deployment page makes all completed live-run papers available, only a small subset was additionally submitted to arXiv. Before submission, each selected paper underwent a minimal human integrity review covering factual accuracy, citation validity, artifact consistency, safety-relevant claims, and explicit AI-generation disclosure; the research idea, experiments, reported results, and claimed contribution were left unchanged. These safeguards were informed by recent analyses of research-integrity risks in AI-generated scholarship~\citep{gupta2025plagiarism,zhu2025safescientist,resnik2026autonomous}. Appendix~\ref{app:arxiv_submissions} lists the arXiv-submitted subset.

%% file: sections/experiments.tex
\section{Human Review Evaluation}
\label{sec:experiments}

Assessing whether FARS produces genuine research contributions requires evaluation that goes beyond automated metrics. A growing number of publicly available AI-based reviewing systems can generate surface-level feedback, but they rarely deliver the deep, expert judgment needed to assess the hypothesis quality, experimental rigor, and actual contribution of a paper. The problem is sharper for AI-generated papers, which can additionally contain fabricated or hallucinated content such as nonexistent citations, unverifiable results, or methods that diverge from their implementation. Current automated reviewers are not designed to detect these issues. We therefore recruit qualified human experts to provide deep and comprehensive reviews of FARS-generated papers, and pair their judgment with an integrity audit grounded in each paper's source artifacts. This section describes the review protocol and reports what the resulting reviews reveal about the quality and integrity of the FARS-generated papers.

\subsection{Review Protocol}
\label{sec:review_protocol}

\paragraph{Reviewer Recruitment} We posted an open call for volunteer reviewers across several social media platforms and collected volunteer information through a questionnaire that recorded each applicant's affiliation, academic title, country, primary research areas, and prior peer-review experience. The call attracted 583 applications, of which 503 (86\%) reported prior reviewing experience, yielding a large and experienced pool from which to select.

\paragraph{Reviewer Screening} From this pool we selected 88 reviewers, screening primarily on three criteria: the match between an applicant's research areas and the topic distribution of the generated papers (\Cref{sec:deployment}), academic standing, and prior peer-review experience. All 88 selected reviewers had reviewed before, and roughly 82\% (72/88) had reviewed more than 20 papers. The panel spans career stages---predominantly PhD students, alongside postdoctoral researchers, faculty, and industry research scientists---and multiple countries, with the largest contingents from China and the United States. Because the reviewers are recruited volunteers rather than an independent venue program committee, the evaluation should be read as a structured expert analysis rather than an independent conference peer-review process.

\paragraph{Review System and Standards} We built a dedicated review system together with a review standard tailored to AI-generated papers,\footnote{\url{https://docs.google.com/document/d/1Lebt2dUVR4igwKO1TeFZVgTA0_2aDDXRB-hEyBNzgpM/edit?usp=sharing}} modeled on the ICLR review form but adapted to FARS's hypothesis-and-validation outputs. The standard retains the ICLR scoring dimensions: overall rating, soundness, presentation, contribution, and reviewer confidence, and adds FARS-specific principles: short papers are not penalized for length, non-standard structure is permitted, and a well-motivated hypothesis that is rigorously falsified is treated as a first-class contribution rather than a failed paper. Because AI-generated papers cannot be reviewed under the usual assumption of good faith, the standard further introduces an \emph{AI Integrity Audit}. To support it, reviewers are given access to each paper's full artifact trace: code repository, experiment logs, and raw output data.  They are required to verify the manuscript's claims against these sources along eight integrity failure modes: hallucinated citations, fabricated or unverifiable results, spurious novelty, hallucinated methods or baselines, mathematical and logical errors, internal inconsistencies, experimental-design pathologies, and language pathologies. Details for each mode are provided in \Cref{app:failure_modes}. 

\paragraph{Review Form and Scoring} Each review follows a fixed structure: summary, strengths, weaknesses, optional suggestions, an ethics and reproducibility assessment, the integrity audit, and a mandatory disclosure of any LLM assistance used while writing the review. The scoring is aligned with ICLR 2026: an overall rating on the discrete scale $\{0, 2, 4, 6, 8, 10\}$, ranging from strong reject (0) to strong accept (10) with 6 marking the acceptance threshold, together with sub-scores for soundness, presentation, and contribution on a 1--4 scale and reviewer confidence on a 1--5 scale; integrity issues are flagged through a separate field. While the numeric scale is identical to ICLR, the rubric is calibrated for short, focused papers, so that a high score does not require exhaustive long-paper breadth but does require a clear, bounded central claim supported by the available evidence.

\paragraph{Assignment} Given the total number of papers and the size of the reviewer panel, each reviewer was asked to review roughly 5--10 papers. Rather than assigning papers centrally, we let reviewers self-claim papers on the review system according to their own research areas and interests, with each paper claimable by at most three reviewers.

\subsection{Quantitative Review Analysis}
\label{sec:quant_analysis}

\paragraph{Reviewer Participation} The review cycle ran from March 21 to April 12, 2026. Before analysis, we audited the 284 submitted reviews and found three paper-review mismatches: one review assigned to FA0027 discussed FA0045 instead, one review assigned to FA0045 duplicated an FA0020 review, and one review assigned to FA0297 referred to a paper outside FARS's papers. We corrected these mismatches by reassigning the FA0027 review to FA0045 and excluding the two invalid reviews, leaving 282 valid reviews for analysis. Of the 88 selected reviewers, 55 submitted at least one review, together producing the 282 valid reviews analyzed here across 140 of the 166 generated papers: 45 papers (27.1\%) received one review, 48 (28.9\%) received two, and 47 (28.3\%) received three, while the remaining 26 (15.7\%) papers were not reviewed and are excluded from all statistics below. 

\begin{figure}[t]
\centering
\includegraphics[width=\textwidth]{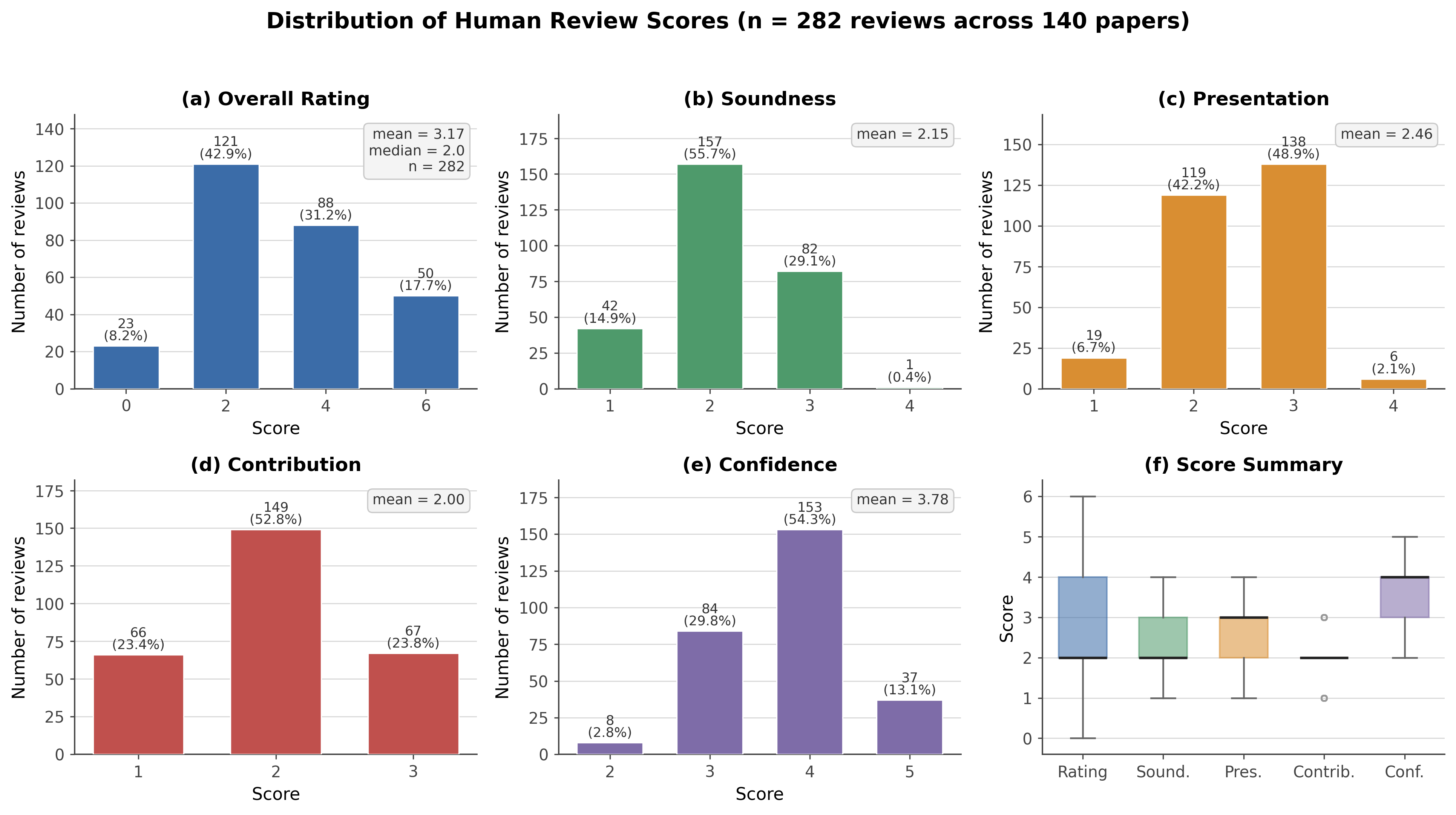}
\caption{Distribution of human review scores across 282 reviews of 140 FARS-generated papers: (a) overall rating, (b--e) sub-scores for soundness, presentation, contribution, and reviewer confidence, and (f) a box-plot summary across all dimensions.}
\label{fig:rating_dist}
\end{figure}

\begin{figure}[t]
\centering
\includegraphics[width=0.62\textwidth]{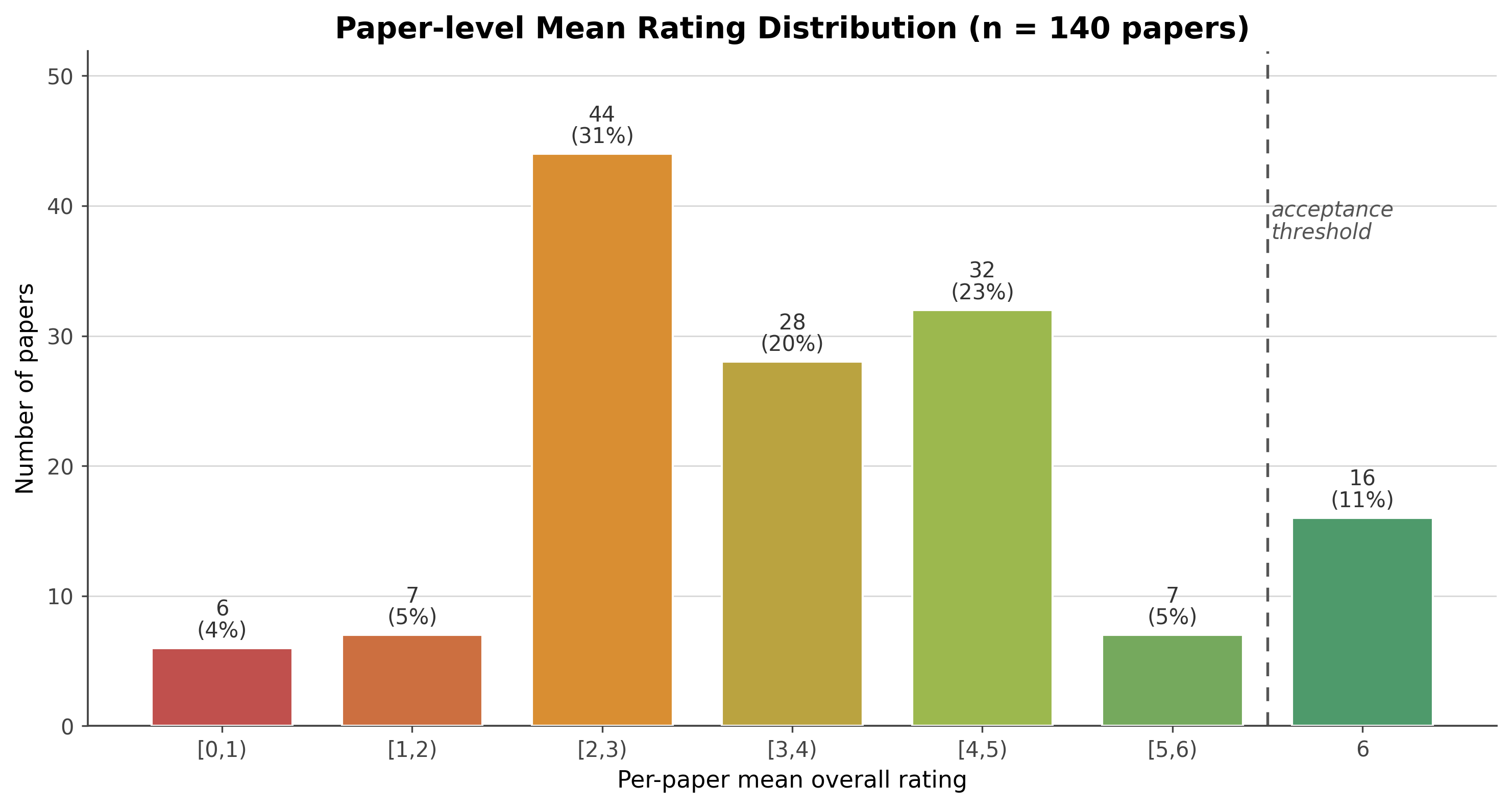}
\caption{Paper-level distribution of mean overall rating across the 140 reviewed papers, binned by integer-width intervals (rightmost bar is exactly 6). }
\label{fig:paper_rating}
\end{figure}

\paragraph{Overall Rating Distribution} Figure~\ref{fig:rating_dist} reports the score distribution over all 282 reviews. The overall ratings are left-skewed and concentrated below the acceptance threshold, with a mean of 3.17 and a median of 2.0. At the review level, 8.2\% of reviews assign a rating of 0, 42.9\% a 2, 31.2\% a 4, and 17.7\% a 6; no review assigned an 8 or a 10, so 6 is the highest score observed in the corpus. Although roughly half of all reviews (48.9\%) place a paper at 4 or above, only 17.7\% reach the ICLR acceptance threshold of 6. Aggregating reviews to the paper level gives a more conservative picture (Figure~\ref{fig:paper_rating}): among the 140 reviewed papers (mean 3.23, median 3.0), 39.3\% attain a mean rating of at least 4, 31.4\% are rated $\geq$4 by every reviewer, and 69.3\% receive at least one rating of 4 or higher, yet only 11.4\% (16 papers) reach a mean rating at or above the acceptance threshold. This last proportion should be interpreted with care, as it is inflated by papers with sparse reviews: 14 of these 16 papers were assessed by a single reviewer, and only 2 of the 95 papers with two or more reviews attain a mean rating of 6. Overall, the corpus exhibits substantial variation in assessed quality alongside a comparatively thin tail of strong outputs.

\paragraph{Sub-score Distributions} The four sub-scores (Figure~\ref{fig:rating_dist}b--e) show where quality concentrates. Soundness (mean 2.15) and contribution (mean 2.00) sit at the lower end. More than half of reviews assign contribution a 2 (52.8\%) and only 23.8\% assign a 3. By contrast, presentation is comparatively strong (mean 2.46), with 48.9\% of reviews assigning a 3. Reviewer confidence is high (mean 3.78) and clusters at 4 (54.3\%), indicating that reviewers were generally comfortable with their assessments. Top sub-scores are rare: contribution never reaches 4, and soundness and presentation reach 4 in only 1 and 6 reviews, respectively.

\begin{figure}[t]
\centering
\includegraphics[width=\textwidth]{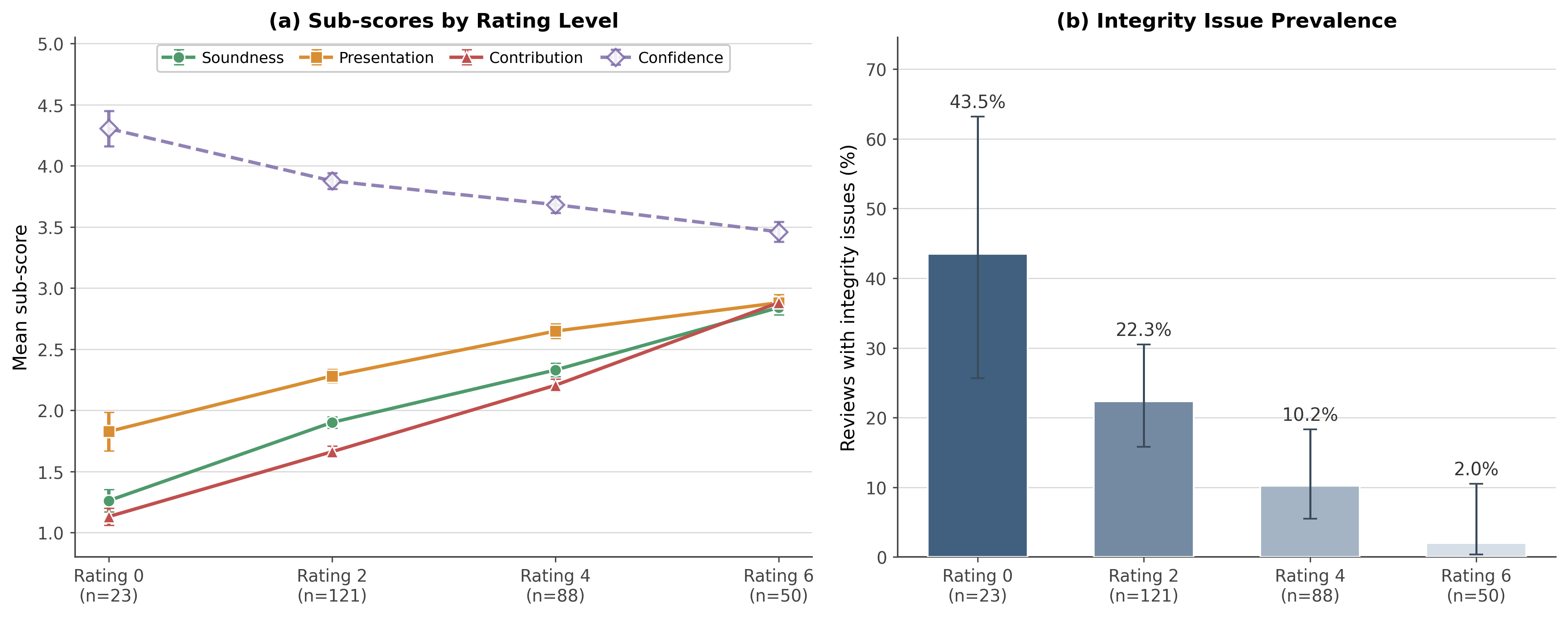}
\caption{Quality stratification and integrity findings for FARS papers. (a) Sub-score means by rating tier show monotonic improvement across soundness, presentation, and contribution, while confidence decreases for higher ratings. (b) Integrity issue prevalence shows a strong inverse relationship with quality, from 43.5\% at rating 0 to 2.0\% at rating 6. Error bars in (a) denote standard errors of the mean; bars in (b) show point estimates with 95\% Wilson confidence intervals; the number of reviews $n$ in each rating tier is annotated on the horizontal axis.}
\label{fig:quality_pred}
\end{figure}

\paragraph{Score Predictors} To identify which dimensions track overall quality, we compute Pearson correlations between each quality sub-score and the overall rating (Table~\ref{tab:correlations}) and compare sub-score means across rating tiers (Figure~\ref{fig:quality_pred}a). Contribution is the strongest predictor ($r=0.743$), followed by soundness ($r=0.652$) and presentation ($r=0.451$), all highly significant. This ordering suggests that the main bottleneck is not manuscript polish but whether a paper delivers a meaningful, well-supported contribution. Reviewer confidence measures certainty rather than quality, so we do not treat it as a predictor. We note that confidence runs counter to rating, being higher on low-rated papers than on high-rated ones (mean 4.30 at rating 0 vs.\ 3.46 at rating 6; Figure~\ref{fig:quality_pred}a), indicating that reviewers are more certain when identifying weak work than when endorsing strong work.

\begin{table}[t]
\centering
\caption{Pearson correlations between the three quality sub-scores and overall rating. Contribution shows the strongest association with overall quality.}
\label{tab:correlations}
\adjustbox{max width=\textwidth}{
\begin{tabular}{lccc}
\toprule
Dimension & Pearson $r$ & $p$-value & Direction \\
\midrule
Contribution & \textbf{0.743} & $1.14 \times 10^{-50}$ & Strong positive \\
Soundness & 0.652 & $1.38 \times 10^{-35}$ & Strong positive \\
Presentation & 0.451 & $1.62 \times 10^{-15}$ & Moderate positive \\
\bottomrule
\end{tabular}
}
\end{table}

\paragraph{Integrity Audit Outcomes} The AI Integrity Audit flagged problems in 47 of 282 reviews (16.7\%), affecting 39 of 140 reviewed papers (27.9\%). These flags concentrate almost entirely in low-rated papers (Figure~\ref{fig:quality_pred}b): 43.5\% of rating-0 reviews report integrity issues, falling to 22.3\% at rating 2, 10.2\% at rating 4, and just 2.0\% at rating 6. The steep gradient indicates that integrity violations behave as a distinct failure mode concentrated in the weakest outputs rather than a smoothly varying quality dimension, consistent with benchmark evidence that open-ended ML research agents can generate plausible papers while struggling to validate experimental results~\citep{chen2025mlrbenchevaluatingaiagents}.

\paragraph{Reviewer Agreement} For the 95 papers with two or more reviews, we quantify how far reviewers' overall ratings diverge using the per-paper spread (highest minus lowest rating). Since adjacent rating levels differ by two points, a spread of 2 corresponds to one level of disagreement. Reviewers assign identical ratings for 28.4\% of these papers, differ by a single level (spread of 2) for 47.4\%, by two levels (spread of 4) for 20.0\%, and by the maximum possible spread of 6 (e.g., 0 versus 6) for 4.2\%. The mean spread is 2.00 points---about one rating level. This degree of divergence is in line with the substantial reviewer disagreement documented for top machine-learning venues: in the NeurIPS 2021 consistency experiment, the average paper scores of two independent committees correlated at only $r\approx0.58$ and the committees disagreed on accept/reject for 23\% of papers~\citep{beygelzimer2023machinelearningreviewprocess}, while a large-scale analysis of ICLR 2024/2025 reports a per-paper reviewer disagreement of roughly $1.5$ points (mean pairwise score difference before rebuttal) on a comparable ten-point scale~\citep{kargaran2025iclrinsights}. Our spread is thus of the same order, situating the FARS review within known peer-review variability rather than indicating anomalously noisy judgments. Because the reviewers are recruited volunteers rather than an independent venue program committee, these agreement statistics should be read as evidence of structured-review consistency for this deployment rather than as a substitute for independent venue review.

\paragraph{LLM-Assisted Reviewing} In their mandatory disclosures, 74.5\% of reviews (210/282) reported using an LLM while preparing the review (GPT/ChatGPT: 81, Claude: 40, Gemini: 30), predominantly for language polishing and summarization rather than substantive evaluation. We record this because LLM-assisted reviewing may influence phrasing, consistency, and attention allocation even when the final judgments and scores remain human-authored.

\subsection{Qualitative Findings}
\label{sec:findings}
The scores in \Cref{sec:quant_analysis} say how good the papers are; the free text of the reviews says \emph{why}. To characterize what reviewers reward and penalize, we label every review's strengths and weaknesses against the category taxonomy of \citet{kargaran2025iclrinsights} and map reported integrity problems onto the eight integrity failure modes of the AI Integrity Audit (\Cref{sec:review_protocol,app:failure_modes}), using a large language model annotator (Claude Opus 4.6, temperature 0). Each strength or weakness point is assigned a main category and subcategory; the percentages below are the share of the 282 reviews whose text was assigned a given category, and a single review may map to several.

\paragraph{What Reviewers Reward} Strengths cluster on execution rather than originality. The most frequently praised dimensions are methodological soundness (63.5\% of reviews), experimental design (61.0\%), and clear motivation (57.4\%), followed by results (48.9\%), while novelty is rewarded least often (31.9\%). At the subcategory level the dominant strengths are a well-motivated problem (48.2\%), an elegant and efficient method (29.1\%), thorough ablation studies (25.2\%), and a clear and correct interpretation of results (23.4\%). Notably, 18.1\% of reviews credit a detailed error or failure analysis---a direct consequence of the FARS design choice to surface negative and null results rather than suppress them. The praise profile sharpens with quality (\Cref{tab:sw_categories}): recognition of novelty rises monotonically from 21.5\% of low-rated reviews to 54.0\% of high-rated ones, with the same upward trend for experiments (52.8\%$\to$78.0\%) and results (38.9\%$\to$64.0\%), indicating that what distinguishes a strong FARS paper is not competent execution alone but a contribution reviewers judge to be genuinely new.

\paragraph{Recurring Weaknesses} Criticism is dominated by evidence quality. Experimental weaknesses appear in 92.9\% of reviews and are essentially universal among the highest-rated papers (100\%), confirming that even the best FARS outputs leave reviewers wanting stronger evidence. The leading complaints are narrow evaluation---too few datasets or limited domain coverage (46.8\%) and poor generalization across settings (46.8\%)---followed by missing ablations (31.9\%) and weak baselines (30.1\%). Part of this concentration reflects format rather than outright failure: the current FARS is configured to produce short, focused papers, whose experimental sections are necessarily less extensive than those of full-length submissions, and the review rubric was calibrated so that brevity alone is not penalized (\Cref{sec:review_protocol}). Even under this calibration, however, experimental concerns remain the single most prevalent weakness, indicating that the problem is often not breadth per se but \emph{sufficiency}---a number of papers fail to support their central claim even with the small but well-targeted set of experiments that a short paper warrants. Beyond experiments, reviewers frequently question methodological soundness (63.1\%; chiefly weak theoretical justification, 27.3\%, and unrealistic assumptions, 22.3\%) and results (58.2\%; contradictory or unexplained findings, 22.3\%, marginal gains, 21.3\%, and overinterpretation, 17.7\%). Weaknesses follow the opposite quality gradient to strengths: novelty criticism falls from 42.4\% of low-rated reviews to 18.0\% of high-rated ones and writing complaints from 24.3\% to 12.0\%, while experimental criticism stays high across all tiers (\Cref{tab:sw_categories}). This pattern---low-rated papers penalized for experimental and methodological flaws, high-rated papers set apart by novelty and soundness---mirrors the rating-tier gradient reported for human-written ICLR submissions~\citep{kargaran2025iclrinsights}, placing FARS outputs within familiar review dynamics rather than a separate regime. It also echoes the quantitative scores: FARS reliably produces a focused, well-motivated claim, but its supporting evidence is too narrow for reviewers to consider the result established.

\begin{table}[t]
\centering
\caption{Reviewer-identified strengths and weaknesses by category, as a percentage of the 282 reviews whose text was assigned each category (multi-label). \emph{All} is the overall rate; \emph{Lo}/\emph{Hi} restrict to reviews with overall rating $\le 2$ ($n{=}144$) and $\ge 6$ ($n{=}50$). Strengths rise and weaknesses fall with rating, except experimental concerns, which persist across all tiers.}
\label{tab:sw_categories}
\begin{tabular}{lcccccc}
\toprule
& \multicolumn{3}{c}{Strength (\%)} & \multicolumn{3}{c}{Weakness (\%)} \\
\cmidrule(lr){2-4}\cmidrule(lr){5-7}
Category & All & Lo & Hi & All & Lo & Hi \\
\midrule
Experiments \& Evaluation         & 61.0 & 52.8 & 78.0 & 92.9 & 88.2 & 100.0 \\
Methodology \& Soundness          & 63.5 & 59.7 & 74.0 & 63.1 & 61.8 & 62.0 \\
Results                           & 48.9 & 38.9 & 64.0 & 58.2 & 60.4 & 48.0 \\
Novelty \& Contribution           & 31.9 & 21.5 & 54.0 & 37.6 & 42.4 & 18.0 \\
Motivation                        & 57.4 & 52.1 & 64.0 &  8.9 & 11.1 &  2.0 \\
Writing \& Presentation           & 38.3 & 40.3 & 36.0 & 20.2 & 24.3 & 12.0 \\
\bottomrule
\end{tabular}
\end{table}

\paragraph{Integrity Failure Modes} The AI Integrity Audit exposes problems that ordinary peer review cannot, because reviewers cross-checked every claim against each paper's code, logs, and raw outputs. Whereas 16.7\% of reviews recorded a \emph{formal} integrity violation (\Cref{sec:quant_analysis}), free-text analysis shows that 44.0\% (124/282) describe at least one integrity failure mode (\Cref{tab:failure_modes}). The most common are experimental-design pathologies (22.7\%; e.g.\ confounded comparisons or suppressed unfavorable runs) and internal-consistency failures (13.1\%; contradictory numbers across abstract, tables, and figures). Crucially, the failure modes differ sharply in how often a mention is escalated to a formal violation: hallucinated methods or baselines (20 of 22 mentions flagged), hallucinated citations (13/16), and fabricated results (16/21) are almost always treated as genuine integrity breaches, whereas language and reasoning pathologies (4 of 31) and many design concerns are recorded as weaknesses rather than violations---precisely the distinction the audit asks reviewers to draw between fabrication and ordinary methodological limitation. That the gravest, code-verifiable failures (code--paper mismatch, untraceable numbers, nonexistent references) are both present and reliably caught underscores why artifact-grounded human review remains necessary for deployment-scale automated research: these are exactly the errors that plausible-looking generated papers conceal and that score- or text-only evaluation would miss~\citep{chen2025mlrbenchevaluatingaiagents}.

\begin{table}[t]
\centering
\caption{Integrity failure modes across the 282 reviews. "Mentioned" counts reviews whose weakness or audit text describes the failure mode; "Flagged" counts how many of those reviews also recorded a formal integrity violation. Severe, code-verifiable failures are almost always escalated to violations, while softer pathologies are usually noted only as weaknesses.}
\label{tab:failure_modes}
\begin{tabular}{lcc}
\toprule
Failure mode & Mentioned (\%) & Flagged \\
\midrule
Experimental-design pathologies   & 64 (22.7) & 22 \\
Internal-consistency failures     & 37 (13.1) & 19 \\
Language / reasoning pathologies  & 31 (11.0) & 4 \\
Hallucinated methods / baselines  & 22 (7.8)  & 20 \\
Fabricated experimental results   & 21 (7.4)  & 16 \\
Hallucinated citations            & 16 (5.7)  & 13 \\
Mathematical / logical errors     & 7 (2.5)   & 3 \\
Spurious novelty / contamination  & 5 (1.8)   & 1 \\
\bottomrule
\end{tabular}
\end{table}

Taken together, the qualitative analysis converges with the quantitative scores: reviewers consistently reward FARS for well-motivated problems and competent execution, the dominant limitation is the sufficiency of experimental evidence rather than presentation, and the artifact-grounded audit reliably surfaces the code-verifiable failure modes that distinguish AI-generated papers from human-written ones. \Cref{tab:representative_outputs} lists representative reviewed papers that illustrate these patterns end to end, from a rewarded falsifiable negative result (FA0328) and a controlled compression study that reveals an accuracy--perplexity tradeoff (FA0029) to a well-posed paper that one reviewer flagged for artifact-level integrity concerns (FA0064). To support scrutiny and follow-up analysis, we release all reviewed papers together with their reviews\footnote{\url{https://huggingface.co/datasets/analemma-ai/fars-a-reviews}}. Before releasing, we contacted all reviewers who participated in the review process, and since some of them preferred not to open-source their own reviews, we withheld these reviews from the public release; this affects six reviews, one each from FA0004, FA0028, FA0050, FA0052, FA0145, and FA0194, while all other reviews for these papers remain available.

\begin{table}[t]
\centering
\caption{Representative reviewed FARS outputs from the public deployment. Ratings are individual overall review scores on the ICLR scale $\{0,2,4,6,8,10\}$; S/P/C report mean soundness, presentation, and contribution sub-scores.}
\label{tab:representative_outputs}
\adjustbox{max width=\textwidth}{
\begin{tabular}{p{0.08\textwidth}p{0.31\textwidth}p{0.12\textwidth}p{0.12\textwidth}p{0.28\textwidth}p{0.07\textwidth}}
\toprule
ID & Title & Ratings & S/P/C & Reviewer takeaway & Links \\
\midrule
FA0328 & Position Bias Correction is Insufficient for One-Pass Attention Sorting & 6, 6, 4 & 2.00 / 2.67 / 2.67 & Reviewers praised the falsifiable negative result and crisp evidence that debiasing alone does not explain iterative Attention Sorting gains. & \href{https://gitlab.com/fars-a/calib-attnsort-onepass/-/blob/master/writing/paper/main.pdf}{PDF} / \href{https://gitlab.com/fars-a/calib-attnsort-onepass}{code} \\
FA0029 & Output-Space Allocation Costs for Calibration-Guided LLM Compression & 6, 6, 4 & 2.67 / 3.00 / 3.00 & Reviewers praised the clean single-change test of ROCKET's allocation cost and the candid analysis of a modest accuracy gain, higher perplexity, and high error-correlation limits. & \href{https://gitlab.com/fars-a/rocket-activation-aware-knapsack/-/blob/master/writing/paper/main.pdf}{PDF} / \href{https://gitlab.com/fars-a/rocket-activation-aware-knapsack}{code} \\
FA0064 & NLL-Guided Full-Attention Layer Selection for Training-Free Sliding-Window Adaptation & 6, 6, 4 & 2.33 / 3.00 / 2.67 & Reviewers praised the clear hypothesis and practical training-free selection rule, while one review flagged artifact-level integrity concerns. & \href{https://gitlab.com/fars-a/nll-guided-swaa-layer-selection/-/blob/master/writing/paper/main.pdf}{PDF} / \href{https://gitlab.com/fars-a/nll-guided-swaa-layer-selection}{code} \\
\bottomrule
\end{tabular}
}
\end{table}

%% file: sections/comparison.tex
\section{Comparison with Prior Research Systems}
\label{sec:comparison}

The human evaluation in \Cref{sec:experiments} characterizes the quality of FARS outputs in isolation; it does not place them relative to other autonomous research systems. A like-for-like human comparison is impractical, because the human panel reviewed only FARS papers and the prior systems release only small numbers of generated papers. To obtain a common yardstick for paper quality, we instead apply a single automated reviewer, the Stanford Agentic Reviewer (SAR)\footnote{\url{https://paperreview.ai/}}, uniformly to the public paper outputs of every system, including FARS. Concretely, we collect the generated papers released by six prior end-to-end research systems: The AI Scientist~\citep{lu2024aiscientistfullyautomated}\footnote{The AI Scientist releases its generated papers in a public Google Drive grouped by base foundation model (Claude~3.5 Sonnet, GPT-4o, DeepSeek Coder, and Llama-3.1 405B). We use only the Claude~3.5 Sonnet subset, which the authors identify as their highest-quality outputs and explicitly recommend for qualitative inspection.}, The AI Scientist-v2~\citep{yamada2025aiscientistv2workshoplevelautomated}, CycleResearcher~\citep{weng2025cycleresearcher}, DeepScientist~\citep{weng2025deepscientist}, AutoResearchClaw~\citep{liu2026autoresearchclawselfreinforcingautonomousresearch}, and ScientistOne~\citep{meng2026scientistonechainofevidence}. We then score these papers, together with the FARS-generated papers, using this same reviewer under an ICLR-style rubric that returns an overall rating on a $0$--$10$ scale. Because the identical reviewer and rubric are applied to all systems, the resulting scores are comparable across systems even though their absolute calibration differs from human review. The Stanford Agentic Reviewer may fail to produce a review. Issues arise when papers exceed the input size limit or errors occur during review generation. We present statistics exclusively for papers that receive valid reviews across all systems including FARS. Accordingly, the paper counts in \Cref{tab:cross_system} represent only successfully reviewed subsets, not the full set of released papers from each system.

\paragraph{FARS attains the highest automated rating} Under this shared reviewer (\Cref{tab:cross_system}), FARS achieves the highest mean rating, $5.00$ over $165$ papers, ahead of every prior system: the strongest baselines, DeepScientist ($4.13$) and ScientistOne ($4.00$), trail by roughly one point, and the pooled mean of all $110$ non-FARS papers is $2.74$, so FARS leads the field by about $+2.3$ points. FARS is also the only system with papers reaching the automated acceptance threshold ($1.8\%$ scored $\geq 6$), whereas no baseline paper does. This advantage holds despite FARS being scored on by far the largest and least curated corpus, whereas several baselines contribute only a handful of papers.

\paragraph{Automated scores are a relative signal, not an acceptance rate} The automated reviewer is on average more lenient than the expert panel: on the same FARS corpus it assigns a mean of $5.00$, against the human mean of $3.23$ (\Cref{sec:experiments}). The absolute automated scores should therefore be read as a relative ranking across systems rather than as a probability of acceptance, and the stricter human evaluation in \Cref{sec:experiments} remains the more conservative measure of FARS quality. Reassuringly, the two measures agree in direction: FARS leads under the automated reviewer, and its human ratings exceed the automated scores that most prior systems obtain.

\paragraph{Scope and caveats} This comparison is for reference only and does not constitute a controlled benchmark. The baseline corpora have small and unbalanced sample sizes ranging from 3 to 71. Calculated means for individual systems contain noise, especially those with just three papers. The systems address diverse problem domains and benchmarks rather than a consistent task set, and each adopted its own criteria for paper release. Finally, the automated reviewer is itself an LLM-based proxy with its own biases. We therefore treat \Cref{tab:cross_system} as evidence that FARS outputs are competitive with or stronger than those of prior systems under a uniform automated standard, while relying on the human study for the more demanding, integrity-aware assessment.

\begin{table}[t]
\centering
\caption{Cross-system comparison under the Stanford Agentic Reviewer, applied uniformly to each system's public paper outputs. Ratings are on a $0$--$10$ scale; \emph{$\geq 6$} is the share of papers reaching the automated acceptance threshold. FARS additionally underwent expert human review (last row, from \Cref{sec:experiments}), shown for calibration; the human rating is markedly stricter than the automated one.}
\label{tab:cross_system}
\begin{tabular}{lcccc}
\toprule
System & Reviewer & \# Papers & Mean rating & $\geq 6$ (\%) \\
\midrule
The AI Scientist~\citep{lu2024aiscientistfullyautomated}                       & SAR & 71  & 2.29 & 0.0 \\
The AI Scientist-v2~\citep{yamada2025aiscientistv2workshoplevelautomated}      & SAR & 3   & 2.40 & 0.0 \\
CycleResearcher~\citep{weng2025cycleresearcher}                                & SAR & 6   & 2.78 & 0.0 \\
AutoResearchClaw~\citep{liu2026autoresearchclawselfreinforcingautonomousresearch} & SAR & 7 & 3.19 & 0.0 \\
ScientistOne~\citep{meng2026scientistonechainofevidence}                       & SAR & 20  & 4.00 & 0.0 \\
DeepScientist~\citep{weng2025deepscientist}                                    & SAR & 3   & 4.13 & 0.0 \\
\textbf{FARS}                                                                  & SAR & 165 & \textbf{5.00} & \textbf{1.8} \\
\midrule
FARS                                                                           & Human     & 140 & 3.23 & 11.4 \\
\bottomrule
\end{tabular}
\end{table}

%% file: sections/conclusion.tex
\section{Conclusion}
\label{sec:conclusion}

We introduced FARS, a fully automated AI-for-AI research system that turns research directions into complete papers through a shared auditable workspace. Its first public deployment produced 166 papers across 67 fine-grained AI/ML topics while preserving the proposals, code, logs, results, and manuscripts behind each output. This corpus lets us evaluate automated research as a deployment-scale distribution rather than a selected set of demonstrations. Across 282 expert reviews of 140 papers, FARS produces reviewable and occasionally strong research artifacts; a uniform automated-review comparison further places FARS competitively relative to prior systems. Together, these results suggest that end-to-end automated research systems can already generate broad, inspectable research output at scale.

The same evidence also defines the limits of current automated research. The strong-output tail remains thin, and the dominant bottlenecks are not surface presentation but contribution, experimental sufficiency, and claim faithfulness. Progress will therefore depend less on producing more plausible manuscripts than on building agents that design decisive experiments, preserve provenance, audit claims against artifacts, learn from failed trajectories, and withhold outputs that are not scientifically supported. FARS is not evidence that automated research is solved; it is an early systems demonstration of what becomes possible when generation, experimentation, provenance, and evaluation are treated as a single auditable process.

%% file: sections/appendix.tex
\section{Deployment Topic Distribution}
\label{sec:appendix_deployment}

\begin{figure}[t]
\centering
\includegraphics[width=\textwidth]{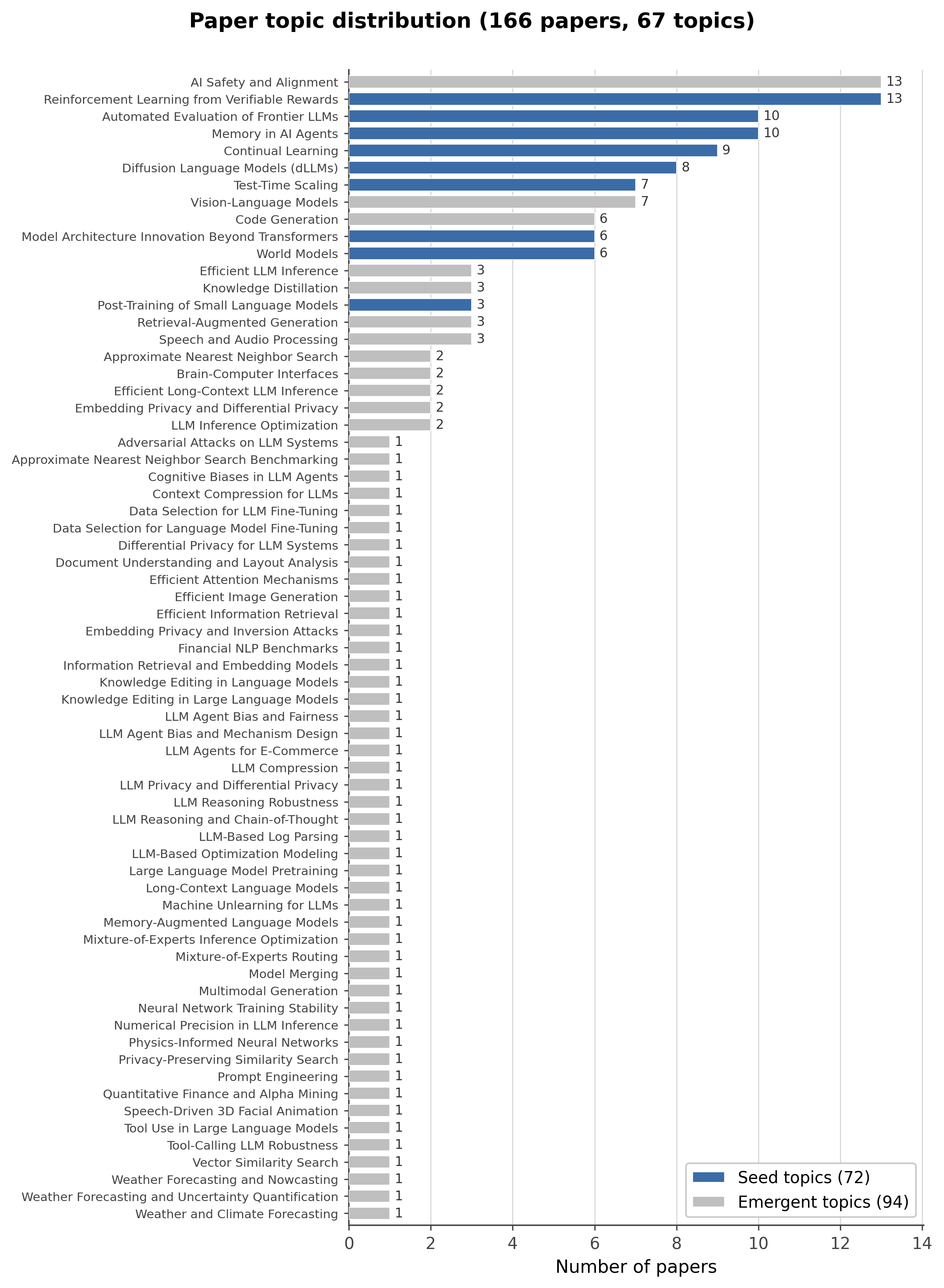}
\caption{Topic distribution of the 166 FARS-generated papers. Blue bars indicate the nine seed topics initially provided to FARS; gray bars indicate emergent topics explored autonomously during deployment.}
\label{fig:topic_dist}
\end{figure}

\Cref{fig:topic_dist} groups the 166 generated papers by whether they fall under the nine seed topics provided at launch or under emergent topics discovered during autonomous exploration.

\section{Integrity Failure Modes}
\label{app:failure_modes}

The AI Integrity Audit in our review standard (\Cref{sec:review_protocol}) asks reviewers to verify each manuscript against its own source artifacts---the code repository, experiment logs, and raw output data---along eight integrity failure modes that AI-generated papers are prone to. We summarize each mode below; the full audit checklist, risk levels, and calibrated worked examples are given in the review standard. A guiding principle throughout is the distinction between a genuine integrity violation and an ordinary methodological weakness: only deceptive or artifact-contradicting cases (fabrication, misattribution, hidden manipulation) are treated as violations, whereas benign metadata artifacts---such as garbled author names for a paper that nonetheless exists, or sub-rounding numerical differences---are noted but not escalated.

\begin{enumerate}
\item \textbf{Hallucinated citations and references.} References whose title, venue, and authors look plausible but that do not exist, or real papers cited in support of a claim they do not actually make. Author-name or year errors for an otherwise correctly identified paper are bibliography artifacts rather than violations.
\item \textbf{Fabricated or unverifiable experimental results.} Reported tables, curves, or statistics that contradict the recorded logs, or that cannot be traced to any artifact at all; also selective reporting of a more favorable configuration than the one the paper describes.
\item \textbf{Spurious novelty and knowledge contamination.} Reproducing established results while claiming novelty, overstating novelty relative to existing work, or asserting a literature gap that does not exist.
\item \textbf{Hallucinated methods and baselines.} Methods or baselines that do not exist in the literature, or whose description in the paper materially diverges from the implemented code (e.g.\ the paper describes one model while the code runs another).
\item \textbf{Mathematical and logical errors.} Proofs or derivations that read fluently but contain material flaws---unstated assumptions, unjustified steps, circular reasoning, or misapplied theorems---that invalidate the claim.
\item \textbf{Internal inconsistencies.} Mutually contradictory statements across the abstract, body, tables, and figures, most often differing numbers reported for the same quantity, sometimes because values from different experiment runs are mixed without disclosure.
\item \textbf{Experimental-design pathologies.} Designs that appear rigorous but bias the outcome: data leakage, compute- or data-unequal comparisons, cherry-picked seeds, post-hoc hypotheses, or suppressed unfavorable runs.
\item \textbf{Language and reasoning pathologies.} Writing patterns that signal shallow understanding---stacked hedging, vacuous generalization, confident incorrectness, narrative in place of evidence, and templated phrasing---which are treated as presentation weaknesses unless coupled with verifiable fabrication from the modes above.
\end{enumerate}

\section{arXiv-Submitted FARS Papers}
\label{app:arxiv_submissions}

\Cref{tab:arxiv_submissions} lists the FARS deployment outputs that were additionally submitted to arXiv. These submissions are provenance-preserving archival versions of the original FARS outputs rather than human-improved research variants: the research ideas, experiments, reported results, and claimed contributions remain those produced during the deployment.

\begin{table}[h]
\centering
\caption{FARS papers submitted to arXiv after minimal human integrity review.}
\label{tab:arxiv_submissions}
\small
\setlength{\tabcolsep}{4pt}
\adjustbox{max width=\textwidth}{
\begin{tabular}{p{0.10\textwidth}>{\raggedright\arraybackslash}p{0.70\textwidth}c}
\toprule
ID & Title & arXiv \\
\midrule
FA0328 & Position Bias Correction is Insufficient for One-Pass Attention Sorting & \href{https://arxiv.org/abs/2606.27793}{2606.27793} \\
FA0064 & NLL-Guided Full-Attention Layer Selection for Training-Free Sliding-Window Adaptation & \href{https://arxiv.org/abs/2606.27791}{2606.27791} \\
FA0029 & Output-Space Allocation Costs for Calibration-Guided LLM Compression: An Empirical Study & \href{https://arxiv.org/abs/2606.27785}{2606.27785} \\
\bottomrule
\end{tabular}
}
\end{table}